# A Semantic-Aware Attention and Visual Shielding Network for Cloth-Changing Person Re-Identification

Zan Gao, *Member, IEEE*, Hongwei Wei, Weili Guan, *Member, IEEE*, Jie Nie, *Member, IEEE*, Meng Wang, *Fellow, IEEE*, and Shengyong Chen, *Senior Member, IEEE*

*Abstract*— Cloth-changing person re-identification (ReID) is a newly emerging research topic that aims to retrieve pedestrians whose clothes are changed. Since the human appearance with different clothes exhibits large variations, it is very difficult for existing approaches to extract discriminative and robust feature representations. Current works mainly focus on body shape or contour sketches, but the human semantic information and the potential consistency of pedestrian features before and after changing clothes are not fully explored or are ignored. To solve these issues, in this work, a novel semantic-aware attention and visual shielding network for cloth-changing person ReID (abbreviated as SAVS) is proposed where the key idea is to shield clues related to the appearance of clothes and only focus on visual semantic information that is not sensitive to view/posture changes. Specifically, a visual semantic encoder is first employed to locate the human body and clothing regions based on human semantic segmentation information. Then, a human semantic attention (HSA) module is proposed to highlight the human semantic information and reweight the visual feature map. In addition, a visual clothes shielding (VCS) module is also designed to extract a more robust feature representation for the cloth-changing task by covering the clothing regions and focusing the model on the visual semantic information unrelated to the clothes. Most importantly, these two modules are jointly explored in an end-to-end unified framework. Extensive experiments demonstrate that the proposed method can significantly outperform state-of-the-art methods, and more robust features can be extracted for cloth-changing persons. Compared with multibiometric unified network (MBUNet) (published in TIP2023), this method can achieve improvements of 17.5% (30.9%) and 8.5% (10.4%) on the LTCC and Celeb-reID datasets in terms of mean average precision (mAP) (rank-1), respectively. When compared with the Swin Transformer (Swin-T), the improvements can reach 28.6% (17.3%), 22.5% (10.0%), 19.5% (10.2%), and 8.6% (10.1%) on the PRCC, LTCC, Celeb, and NKUP datasets in terms of rank-1 (mAP), respectively.

## I. INTRODUCTION

THE person re-identification (ReID) task is to explore the usefulness of image retrieval techniques in the public security domain. It is an upstream task of the person detection or person localization tasks, while the ReID task needs to find additional clues about the target person based on the already-acquired person images or video sequences. A typical person ReID system aims to discover matching persons from a gallery library and return the retrieval sequence based on the query probe. Furthermore, as air pollution continues to rise, individuals frequently wear face masks as a precautionary measure in their everyday routines. Additionally, surveillance cameras often capture face images at a noticeably reduced size. Consequently, even advanced face recognition methods often struggle to accurately identify individuals under such circumstances. To solve this issue, researchers [1], [2], [3], [4], [5], [6], [7], [8], [9], [10], [11], [12], [13], [14], [15], [16], [17] have developed the person ReID technique, which is an important supplement to the face recognition technique [18], [19] and a special case of feature extraction [20], [21], [22], and several person ReID datasets [23], [24], [25] have been released. Moreover, Ye et al. [17] have surveyed the existing person ReID approaches. We can find that these approaches are very effective for the person ReID task with short time spans where the human appearance features are fully used for visual matching, but when the surveillance acquisition period becomes longer, the complexity of clothing changes subsequently increases. Fig. 1 shows some examples of cloth-changing person ReID images, where each row displays the images of the same person wearing different clothes. From them, we can observe that the differences in the visual appearances of the same person with different clothes are

Manuscript received 17 July 2022; revised 9 October 2023; accepted 29 October 2023. This work was supported in part by the National Natural Science Foundation of China under Grant 62372325 and Grant 61872270; in part by the Young Creative Team in universities of Shandong Province under Grant 2020KJN012; in part by the Jinan 20 Projects in universities under Grant 2020GXRC040; and in part by the Shandong Project toward the Integration of Education and Industry under Grant 2022PYI001, Grant 2022PY009, and Grant 2022JBZ01-03. *(Corresponding authors: Hongwei Wei; Zan Gao.)*

Zan Gao is with the Shandong Artificial Intelligence Institute, Qilu University of Technology (Shandong Academy of Sciences), Jinan 250014, China, and also with the Key Laboratory of Computer Vision and System, Ministry of Education, Tianjin University of Technology, Tianjin 300384, China (e-mail: zangaonsh4522@gmail.com).

Hongwei Wei is with the Shandong Artificial Intelligence Institute, Qilu University of Technology (Shandong Academy of Sciences), Jinan 250014, China (e-mail: wmolang@163.com).

Weili Guan is with the Faculty of Information Technology, Monash University, Clayton, VIC 3800, Australia (e-mail: honeyguan@gmail.com).

Jie Nie is with the College of Information Science and Engineering, Ocean University of China, Qingdao 266100, China (e-mail: niejie@ouc.edu.cn).

Meng Wang is with the School of Computer Science and Information Engineering, Hefei University of Technology, Hefei 230009, China (e-mail: eric.mengwang@gmail.com).

Shengyong Chen is with the Key Laboratory of Computer Vision and System, Ministry of Education, Tianjin University of Technology, Tianjin 300384, China (e-mail: sy@ieee.org).

Color versions of one or more figures in this article are available at https://doi.org/10.1109/TNNLS.2023.3329384.

Digital Object Identifier 10.1109/TNNLS.2023.3329384

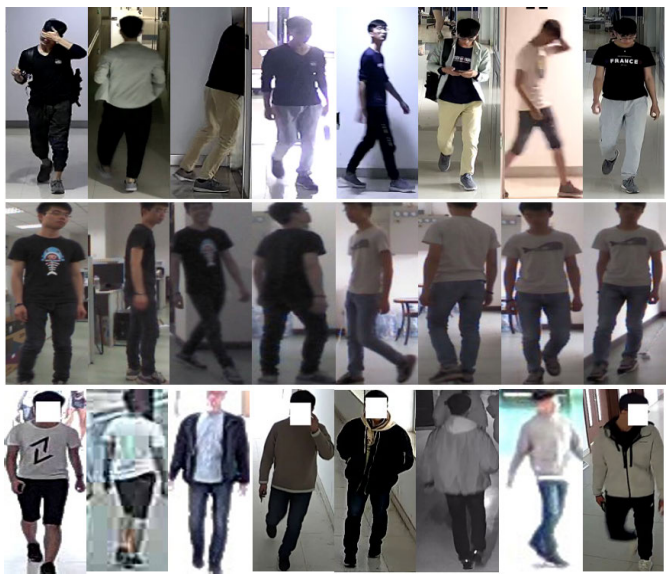

Fig. 1. Examples of cloth-changing person ReID images. The images in each row belong to the same person with different clothes.

very large, and it is also very difficult for humans to identify them. In other words, when data from different cameras are collected for a long time, the appearance of the person's clothing cannot be used as the feature representation. If the existing person ReID approaches are directly applied in this case, their performance deteriorates dramatically, and they often fail. The reason for this is that the above approaches mainly rely on visual clothing appearance, and they cannot provide a robust feature representation for each identity (ID) with different clothes; thus, discriminative and robust feature extraction is an urgent issue for the cloth-changing person ReID task.

To date, a few researchers [26], [27], [28], [29], [30], [31], [32], [33], [34], [35], [36] have made useful attempts for the cloth-changing person ReID task. For example, Huang et al. [27] proposed an augmented representation with vector neurons for cloth-changing. Yang et al. [29] proposed human contour information and polar coordinate transformation to obtain the results of pedestrian matching. Qian et al. [30] proposed a shape embedding module and a clothing-eliminating shape-distillation module (SE + CESD). To enrich clothing styles, Zheng et al. [31] proposed a generative adversarial model (GAM) module to expand the training data. Yu et al. [32] proposed a new solution by involving rich clothing templates in training, and in the retrieval, different clothing templates are added into the query in sequence. Jin et al. [33] proposed a framework called GI-ReID that leverages gait recognition as an auxiliary task to learn cloth-agnostic representations for efficient and latency-free person ID matching in surveillance. Yang et al. [34] proposed SirNet-based on positive and negative sample clustering to increase interclass differences and reduce intraclass gaps. These approaches are very good for trying to solve the cloth-changing person ReID problem, but since the human appearance exhibits large variations with different clothes, it is very difficult for existing approaches to extract discriminative and robust feature representations. Moreover, current works mainly focus on body shape or contour sketches, but the human semantic information and the potential consistency of pedestrian features before and after changing clothes are not fully explored or are ignored.

To solve these issues, in this work, we propose a novel end-to-end SAVS algorithm for the cloth-changing person ReID task to obtain more discriminative and robust features that are irrelevant to clothes. The two key points of the problem to be solved in the field of cloth-changing person ReID: 1) the same person wearing different clothes and 2) different people wearing the same clothes. The SAVS method has already eliminated both effects from appearance simultaneously by attentional weighting and shielding pixels. Extensive experimental results on four cloth-changing person ReID datasets demonstrate that SAVS can outperform state-of-the-art person ReID approaches, and more discriminative and robust features can be obtained that can effectively solve the cloth-changing issue. The main contributions of this article are summarized as follows.

1) We develop a novel end-to-end SAVS network for cloth-changing person ReID that consists of visual semantic encoding and visual semantic decoding. The key idea is to shield clues related to the appearance of clothes and only focus on visual semantic information that is not sensitive to view/posture changes. In this way, the negative effect of the clothing information can be reduced as much as possible.
2) We design a human semantic attention (HSA) module to highlight the human information and reweight the visual feature map that is very helpful for obtaining more discriminative features, and then we develop a visual clothes shielding (VCS) module to extract a more robust feature representation by focusing the model on the visual semantic information unrelated to the clothes. Most importantly, these two modules are jointly explored in an end-to-end unified framework. In this way, more discriminative and robust features can be extracted that are irrelevant to the cloth-changing or pose variants.
3) We systematically and comprehensively evaluate the SAVS algorithm on four public cloth-changing person ReID datasets, and the experimental results demonstrate that the SAVS approach can obtain more discriminative and robust features that are irrelevant to clothes; moreover, it can significantly outperform state-of-the-art cloth-changing person ReID methods in terms of the mean average precision (mAP) and rank-1.

The remainder of this article is organized as follows. Section II introduces the related work, and Section III describes the proposed SAVS method. Section IV describes the experimental settings and the analysis of the results. Section V presents the details of the ablation study, and concluding remarks are presented in Section VI.

## II. RELATED WORK

To date, many person ReID approaches have been proposed. According to the person's visual appearance, these methods can be roughly divided into clothing-consistent person ReID and cloth-changing person ReID. In the following, we will separately introduce them.

### A. Clothing-Consistent Person ReID

In earlier times, people [31], [37], [38] made efforts to develop related methods for clothing-consistent person ReID where the visual appearance of the clothes was consistent for the same person. For example, Sun et al. [9] proposed a part-based convolutional baseline (PCB) module where a base feature map was first obtained and then equally divided into six feature blocks in the horizontal direction. This method is simple but very effective and has become an important benchmark in the field of person ReID. Wang et al. [14] proposed the multiple granularity network (MGN), a multi-branch deep network architecture where one branch is built for the global feature representation and two branches are built for the local feature representation to capture the preference information of the pedestrian ID categories from the whole image. In this way, discriminative information with various granularities can be obtained via an end-to-end feature learning strategy. Gao et al. [16] proposed a deep spatial pyramid-based collaborative feature reconstruction model (DCR) where all blocks of the person were jointly reconstructed; in this way, the issues of occlusion, pose changes, and observation perspective changes can be solved. In addition, some methods employ human skeleton points or human surface texture as a priori knowledge to guide pedestrian ReID. Song et al. [39] proposed a mask-guided contrastive attention model (MGCAM) to learn features separately from the body and background regions; moreover, a novel region-level triplet loss was designed to restrain the features learned from different regions. Miao et al. [40] proposed a novel pose-guided feature alignment (PGFA) method where a pose estimator was utilized to detect key points of the human body in pedestrian images, and then these keypoints were used to decide whether a specific body part was occluded. Gao et al. [5] proposed a novel texture semantic alignment (TSA) approach with visibility awareness for the partial person ReID task, where the occlusion issue and changes in poses were simultaneously explored in an end-to-end unified framework. Zheng et al. [41] proposed for the first time to solve the person ReID problem in 3-D space by learning features from human appearance and 3-D geometric structure in a coherent manner. Wang et al. [15] proposed a novel spatial rescaling (SpaRs) layer to help convolutional neural networks (CNNs) to see more, and it introduced spatial relations among the feature map activations back to guide the model to focus on a broad area in the feature map. Hou et al. [2] proposed a novel interaction-aggregation-update (IAU) block to comprehensively leverage the spatial–temporal context information for high-performance person reID. Zhang et al. [1] introduced a unified attribute-guided collaborative learning scheme tailored for partial person ReID. In their work, they proposed an adaptive threshold-guided masked graph convolutional network, which effectively incorporates human attributes and a cyclic heterogeneous graph convolutional network. This integration facilitates the fusion of cross-modal pedestrian information through both intragraph and intergraph interactions. Ye et al. [17] performed a comprehensive examination of closed-world person ReID from three distinct viewpoints, providing an in-depth analysis. They further evaluated the strengths of existing person ReID approaches and proposed a robust attention generalized mean pooling with weighted triplet loss (AGW) baseline method. Although these models are robust to changes caused by poses, lighting, and viewing angles, they are vulnerable to clothing changes, as the models heavily rely on the consistency of the appearance of clothes.

### B. Cloth-Changing Person ReID

Since the visual appearance of the pedestrians in the cloth-changing person ReID task changes dramatically after a long period, it is very difficult to extract more discriminative and robust feature representations. If the existing clothing-consistent person ReID methods are directly applied to this task, their performance is unsatisfactory. Therefore, to accelerate the development of cloth-changing person ReID techniques, some cloth-changing person ReID datasets have been built and released, such as LTCC [30], PRCC [29], Celeb-reID [27], and NKUP [42]. Moreover, several researchers [27], [28], [29], [30], [35], [36], [43], [44], [45], [46] have made some attempts to address this problem and then assessed their performance on a certain dataset. For example, Yang et al. [29] proposed a spatial polar transformation (SPT) + angle-specific extractor (ASE) module, where human contour sketching information was used to substitute for human color information. Moreover, an SPT layer was designed to transform the contour sketch image, and then a multistream network was used to aggregate multiple granular features to better discriminate people by changing the sampling range of the SPT layer. In this way, the changes in visual appearance caused by clothing changes could be reduced. Qian et al. [30] proposed an SE + CESD module, where the main idea was to completely delete information related to the appearance of clothes and only focus on body shape information that is not sensitive to changes in perspective and posture. The former was used to encode shape information from human body keypoints, and the latter was utilized to adaptively distill the ID-relevant shape features. Huang et al. [27] designed a ReIDCaps module where a vector neuron concept was proposed. For each vector neuron, its direction was used to represent the changes in clothing information, and its length was utilized to identify the people. In this way, the clothing changes of a specific person can be perceived, and the auxiliary modules can be used to enhance the robustness of the module. Zheng et al. [31] proposed a jointly couples discriminative and generative learning in a unified network (DG-Net) module where a generative model was utilized to automatically generate person images with different appearances regarding clothing. Li et al. [26] and Yu et al. [32] proposed a new solution for changing clothes called clothes changing person set (COCAS) and COCAS plus (COCAS+), respectively, where rich clothing templates were supplied; thus, in the query, both the clothing template image and an image of the target person wearing other clothes were fed into the module to find the target image. Gao et al. [45] proposed a novel multigranular visual-semantic embedding algorithm (MVSE) for cloth-changing person ReID, where visual semantic information and human attributes are embedded into the network. Hong et al. [44] proposed a fine-grained shape-appearance mutual learning framework that can learn fine-grained discriminative body shape knowledge in a shaped stream and transfer it to an appearance stream to complement the clothing-unrelated knowledge in the appearance features. Shu et al. [46] proposed a semantic-guided pixel sampling approach for the cloth-changing person ReID task which forces the model to automatically learn clothing-irrelevant cues that are irrelevant to upper clothes and pants. Gu et al. [47] proposed a clothes-based adversarial loss (CAL) to mine clothes irrelevant features from the original RGB images by penalizing the predictive power of the ReID model. Yang et al. [36] proposed a causality-based autointervention model (AIM) to mitigate clothing bias for robust cloth-changing person ReID. Yang et al. [43] proposed an Auxiliary-free Competitive IDentification (ACID) model to achieve a win–win situation by enriching the ID-preserving information conveyed by the appearance and structure features while maintaining holistic efficiency. Zhang et al. [35] proposed a novel multibiometric unified network (MBUNet) for learning the robustness of cloth-changing ReID model by exploiting clothing-independent cues. Since the human appearance with different clothes exhibits large variations, it is very difficult for existing approaches to extract discriminative and robust feature representations, and their performances need to be further improved. Moreover, the human semantic information and the potential consistency of pedestrian features before and after changing clothes are not fully explored or are ignored. Thus, in this work, we fully explore the available visual semantic information and the potential consistency of features and then extract a generalized and robust feature to represent a person wearing different clothes.

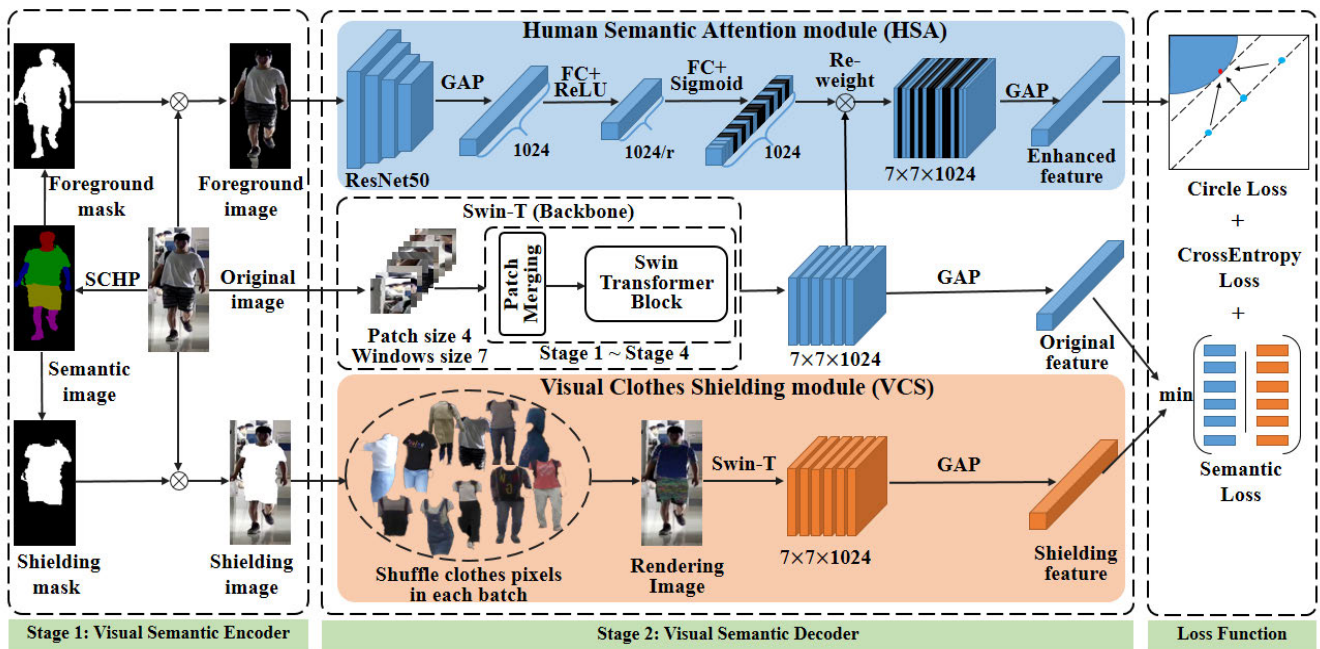

Fig. 2. Pipeline of the SAVS approach is an end-to-end network architecture, and it mainly consists of the visual semantic encoder, the visual semantic decoder, and the loss function. In the visual semantic encoder, the foreground information and visual clothing shielding information are generated for each ID to enrich the image representations. The visual semantic decoder focuses the module on human semantic information and visual clothing shielding information; thus, a more discriminative and robust feature can be extracted. Note that in the SAVS, Swin-T is used as the backbone. GAP indicates the global average pooling, and FC denotes the fully connected layers.

## III. PROPOSED APPROACH

As shown in Fig. 2, our proposed SAVS method consists of two stages: a visual semantic encoder and a visual semantic decoder. Moreover, the visual semantic decoder mainly consists of the backbone, the HSA module, and the VCS module, where these three modules are jointly explored in an end-to-end unified framework. In addition, the loss function is used to guide the network optimization. Specifically, the foreground image of the human body (with background pixel values of 0), the original image, and the shielding image are fed into the HSA, the backbone, and VCS streams, to obtain the enhanced feature, the original feature, and the shielding feature, respectively, and then the loss functions are further utilized to mine their relationship among these features. Since the clothes information is included in the enhanced feature and the original feature, thus, we invalidate the clothing information by semantic loss through the VCS stream, at this point the only human semantic information left is the cues that are not related to the clothing, such as facial features, body shape, posture, accompanying markers, and other comprehensive information. In other words, the idea of extracting human semantic information is to first eliminate the influence of background changes through HSA to highlight all human features, and then eliminate the influence of cloth-changing through VCS to highlight human features unrelated to clothes, i.e., effective human semantic information for cloth-changing person ReID. Notice that "Human semantic information" is summarized as biometric cues that are helpful in identifying people. In the following, we will introduce the visual semantic encoder, the visual semantic decoder, and the loss function. Algorithm 1 shows the complete procedure of the proposed SAVS model.

### A. Visual Semantic Encoder

With the development of deep learning techniques [48], [49], researchers have designed different CNNs for the person ReID task, but since visual appearances are very different in the cloth-changing person ReID task, it is very difficult for an individual feature to extract a generalized and robust feature to represent a person with different clothes. To accommodate clothing variations within the limited data, enriched feature representations for each ID are needed. Thus, in the visual semantic encoder, the foreground image and the shielding image are generated with the help of human semantic segmentation maps. Specifically, for the original image, the pretrained self-correction for human parsing (SCHP) module [50] is employed to obtain the human semantic segmentation information, where the human body is divided into 18 semantic parts. To make it suitable for the cloth-changing person ReID task, these 18 semantic parts are recombined to obtain seven parts, including background, head, torso, pants, arms, legs, and belongings. Fig. 3 shows

**Algorithm 1** Training Procedure of the **SAVS**

---
**Input**: Input image $I \in R^{3 \times H \times W}$
  The corresponding semantic segmentation map $M \in R^{1 \times H \times W}$
  epochs $= T$, $t = 0$, the learning rate $= 3.5e^{-3}$,
  the shrink rate $= 0.1$ at 40 to 60
  initialize the parameters of the swim-transformer(backbone)
**Output**: the SAVS parameters $\theta$

---
**while** $t < T$ **do**
1: Extract the foreground image and the shielding image from $I$
   based on the semantics of the segmentation map $M$
2: Obtain the original feature map $F_o \in R^{7 \times 7 \times 1,024}$
   from $I$ by Swin-Transformer (Swin-T)
3: Obtain the original feature $F'_o \in R^{1,024}$
   by using global average pooling
4: Obtain the enhanced feature $F'_e \in R^{1,024}$
   by using equation (1) and (2) in HSA module
5: Obtain the shielding feature $F'_s \in R^{1,024}$ in VCS module
6: Compute the semantic loss by using equation (5)
   to minimize the distance between $F'_o$ and $F'_s$
7: Compute the cross-entropy classification loss
   and the circle loss using equation (4) based on $F'_e$
8: Train the SAVS coaching by the total loss
   and update the SAVS parameters $\theta$
**end while**

---

the results of the human semantic segmentation maps. In the following, we provide detailed information on how the foreground image and the visual shielding image can be obtained.

*1) Foreground Image:* Many previous works focus on obtaining the global or local features for each cloth-changing person, but in this work, we pay more attention to the latent association between the foreground and background information, and the foreground image is considered another image representation of the original image. Thus, the key step is to separate the foreground and background for each original image. Based on the human semantic segmentation maps, we perform a binarization process to distinguish between background and nonbackground. All information other than the background is used as the foreground information, such as the torso and legs. Thus, the foreground mask can be obtained. Finally, we associate each body part with its corresponding mask. These pixels inside the mask boundary or outside the mask boundary are considered the foreground image or the background image, respectively. Fig. 3 provides the results of the foreground images.

*2) Visual Shielding Image:* In the cloth-changing person ReID scenario, the most common change occurs for the upper clothes and pants. To obtain a more discriminative feature for cloth-changing tasks, the visual shielding image is obtained as another new image representation of the original image by covering the clothing regions of the upper clothes and pants. Specifically, unlike separating foreground information, at this step, finer-grained segmentation labels are required to accurately find the local locations of the upper clothes and pants based on the aforementioned semantic segmentation information. In this way, the shielding mask can be obtained where the pixel value is set to one if it belongs to the mask, or its value is set to zero. Finally, we further combine the original image and the shielding mask by matrix multiplication; thus, the visual shielding image can be obtained. Fig. 3 also displays the results of the visual shielding images.

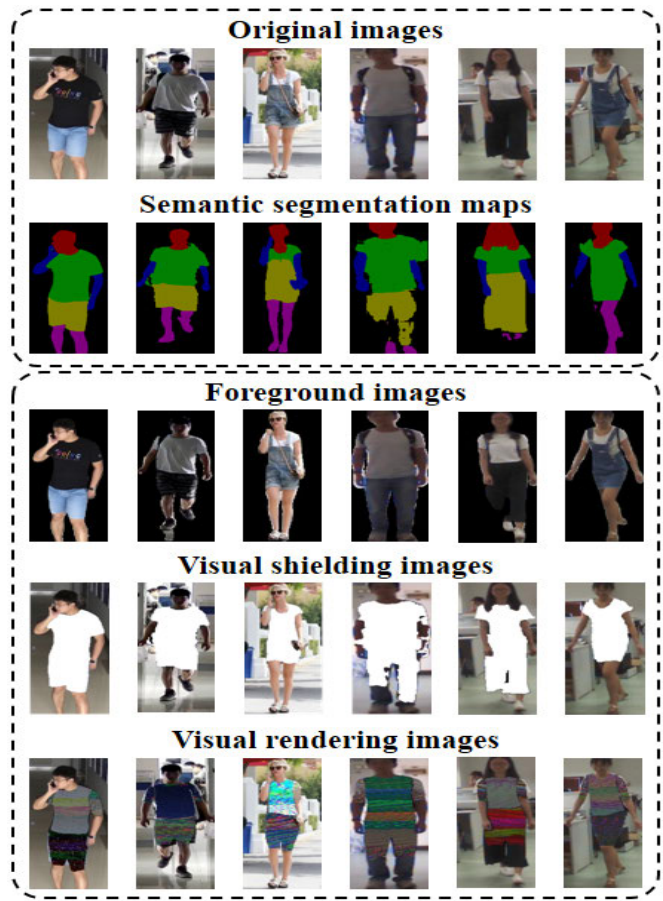

Fig. 3. Results of the visual semantic encoder process. From top to bottom: Original images, the corresponding semantic segmentation maps, the foreground images, the visual shielding images, and the visual rendering images.

### B. Visual Semantic Decoder

The visual semantic decoder mainly consists of the backbone, the HSA module, and the VCS module, where these three modules are jointly explored in an end-to-end unified framework. In our experiments, the Swin Transformer (Swin-T) [51] is used as the backbone to obtain the original feature map $F_o \in R^{7 \times 7 \times 1,024}$, whose input is the original image, and then global average pooling (GAP) is utilized to obtain the original feature $F'_o \in R^{1,024}$. Note that to accommodate person feature extraction, in Swin-T, we set the patch size to 4 for patch partitioning, and we set the window size to 7 for each patch to compute the self-attention inside each window. We keep the transformer blocks from stage 1 to stage 4 to extract visual features and discard the final classification layer of the network. Moreover, the parameters of the Swin-T pretrained on the ImageNet dataset are further used in the following joint optimization. In the following, we will introduce the HSA and VCS modules separately.

*1) Human Semantic Attention Module:* Current works mainly focus on body shape or contour sketches, but human semantic information has not been fully explored. Thus, in this work, an HSA module is designed to highlight the human semantic information and reweight the visual feature map, and the latent correlation between the channels of the convolved features is mined. In this way, the negative effect of the background information can be reduced as much as possible, and more discriminative features can be obtained.

The structure of the HSA module is shown in Fig. 2. In this module, ResNet50 is used as the basic network architecture, and it is first pretrained on the ImageNet dataset. Then, the parameters of ResNet50 are further jointly optimized with other networks. Specifically, in the HSA module, the feature map $F_A \in R^{7 \times 7 \times 1,024}$ is first obtained by ResNet50, and then it can be further fed into the GAP to obtain the feature vector $F'_A \in R^{1024}$. Moreover, the feature vector $F'_A$ is subsequently passed through two fully connected layers, where the first layer is used to obtain a feature representation with reduced dimensionality $(1,024/r)$ (in our experiment, $r$ is set to 16), and the second layer is used to increase the feature dimensionality $(1,024)$. This can be defined as

$$F_w = \sigma\{W_2 \cdot \delta\{W_1 \cdot f_g(F_A)\}\}. \tag{1}$$

$F_w \in R^{1024}$ denotes the weighted feature vector with human semantic information whose input image is the foreground image. $f_g$ indicates the operation of GAP. $W_1$ and $W_2$ represent the parameters of two fully connected layers. $\sigma$ and $\delta$ indicate the sigmoid activation and the rectified linear unit (ReLU) function, respectively. To make the original feature highlight the human semantic information, the reweight operation is further used for the original feature map $F_o \in R^{7 \times 7 \times 1,024}$ obtained from the original image and the backbone (Swin-T) by the weighted feature vector $F_w$, and it can be obtained by

$$F_{e'} = f_g\{F_w \otimes F_o\} \tag{2}$$

where $\otimes$ denotes the channelwise multiplication between the weight vector $F_w$ and the original feature map $F_o$. Moreover, the GAP operation is used to obtain the feature vector that can describe the persons for subsequent classification training. $F'_e \in R^{1,024}$ indicates the enhanced feature, which is the output of the reweight operation and GAP. Since $F_w$ is obtained from the foreground image that can selectively emphasize the human feature channels, $F'_e$ can pay more attention to the human semantic information, and the negative effect of the background information can be reduced as much as possible. Therefore, the extracted feature is more discriminative and robust.

*2) Visual Clothes Shielding Module:* In the cloth-changing person ReID scenario, the clothes of the person often change; thus, it will be very difficult to obtain more discriminative features for cloth-changing tasks. We hope that we can extract clothing-irrelevant features where more attention is given to human semantic information; thus, the VCS module is designed, which focuses the model on visual semantic information unrelated to clothes. Specifically, suppose that there are $b$ original images and $b$ corresponding segmentation maps in each batch of the training stage, which can be denoted as $I = [I_1, I_2, \ldots, I_i, \ldots, I_b]$ and $M = [M_1, M_2, \ldots, M_i, \ldots, M_b]$, respectively. $I_i \in R^{3 \times H \times W}$ and $M_i \in R^{1 \times H \times W}$ are the original image and the corresponding semantic segmentation map, respectively, and $H$ and $W$ separately denote the height and width. According to these segmentation maps, we can obtain the corresponding clothing regions from the original image; moreover, all pixel values of these clothing regions can also be obtained. To confuse these clothes, we build a shielding pixel pool for each batch where all pixels of these clothing regions are shuffled (the ellipse of Fig. 2 shows the pixel pool). Note that in our VCS, we do not care about the pedestrian's upper clothes and pants, and all pixels are equally treated. To reduce the negative effect of the clothes, we specifically transform each pixel of the clothing regions in the original image, and its value is replaced by another value randomly obtained from the shielding pixel pool, but the values of other pixels from nonclothing areas are kept the same as the original image. Finally, the visual rendering image can be obtained, and its results can be observed in the last row of Fig. 3. Then, the visual rendering image is further fed into the Swin-T and GAP to obtain the visual shielding feature $F'_s \in R^{1,024}$, where the network architecture is the same as the backbone of the SAVS, and the shared network parameters with the backbone are employed. Moreover, to make the module focus on the human nonclothing regions, such as the head, face, legs, and feet, we also seek to ensure that the difference between the visual shielding feature and the original feature is as small as possible. In this way, it is difficult for the learning module to differentiate the clothing regions and nonclothing areas, and the clothing-irrelevant features can be extracted, where more attention is devoted to the human semantic information.

### C. Loss Function

The person ReID task is often regarded as a person classification problem; thus, the classification loss is often calculated. To further improve the feature discrimination ability of the proposed method, a metric learning loss is added, and it is used to narrow the intraclass distance and increase the interclass loss. Finally, in the HSA module, the human semantic alignment loss is also utilized. Thus, in total, the loss function of the SAVS can be defined as follows:

$$\mathcal{L} = \lambda_1 \mathcal{L}_{\text{id}} + \lambda_2 \mathcal{L}_{\text{cir}} + \lambda_3 \mathcal{L}_{\text{sem}} \tag{3}$$

where $\mathcal{L}$ is the total loss function of the SAVS, $\mathcal{L}_{\text{id}}$ denotes the classification loss, $\mathcal{L}_{\text{cir}}$ represents the metric learning loss, and $\mathcal{L}_{\text{sem}}$ is the human semantic alignment loss. $\lambda_1$, $\lambda_2$, and $\lambda_3$ are the trade-off parameters for balancing the contributions of each term. In our experiments, each term is equally treated; thus, all $\lambda_1$, $\lambda_2$, and $\lambda_3$ values are set to 1. Specifically, for the classification loss, the public cross-entropy loss is used as the ID loss to learn discriminative features. To make the feature more discriminative, the circle loss [52] in metric learning is employed to measure the distance between sample pairs. Given a single sample $I_e$ in the feature space, let us assume that there are $K$ intraclass similarity scores and $L$ interclass similarity scores associated with $I_e$. Moreover, these intraclass similarity scores and interclass similarity scores are denoted as $s_p^i$ $(i = 1, 2, \ldots, K)$ and $s_n^j$ $(j = 1, 2, \ldots, L)$, respectively. To maximize the intraclass similarity $s_p$ and minimize the interclass similarity, the metric learning loss can be calculated by

$$\mathcal{L}_{\text{cir}} = \log\left[1 + \sum_{i=1}^{K}\sum_{j=1}^{L} \exp\left(\gamma\left(\alpha_n^j s_n^j - \alpha_p^i s_p^i\right)\right)\right]$$
$$\begin{cases} \alpha_p^i = [O_p - s_p^i]_+ \\ \alpha_n^j = [s_n^j - O_n]_+ \end{cases} \tag{4}$$

where $\alpha_n^j$ and $\alpha_p^i$ are nonnegative weighting factors for intraclass similarity scores and interclass similarity scores, respectively. $O_p$ and $O_n$ indicate the optimization score values for $s_p^i$ and $s_n^j$, respectively. $[*]_+$ indicates the optimization process. $\gamma$ is the scale factor where the $\gamma$ values are set to 32 for all similarity scores in our experiments. When a similarity score deviates far from its optimum (i.e., $O_n$ for $s_n^j$ and $O_p$ for $s_p^i$), it should obtain a large weighting factor to obtain an effective update with the large gradient. In this way, it can make the learned feature more discriminative to distinguish between different people wearing similar clothing.

The clothing regions in the visual rendering image are shielded, whose pixels are replaced by the shielding pixel pool and are very different from those of the original image, but the ID information is retained. Thus, to make the module focus on the human nonclothing regions, such as the head, face, legs, and feet, we also seek to ensure that the difference between the visual shielding feature and the original feature is as small as possible. In this way, the clothing-irrelevant features can be extracted, where more attention is given to the human semantic information. Therefore, the mean square error between the visual shielding feature and the original feature is utilized as the human semantic alignment loss, which can be calculated by

$$\mathcal{L}_{\text{sem}} = \frac{1}{b}\sum_{i=1}^{b}(||F_o' - F_s'||_2) \quad (5)$$

where $b$ is the batch size and $||*||_2$ indicates the $L_2$ normalization. $F_o' \in R^{1,024}$ and $F_s' \in R^{1,024}$ denote the original feature and the visual shielding feature, respectively. After optimization, the difference between $F_o'$ and $F_s'$ is very small, and these features focus on the clothing-irrelevant regions of the human. In this way, feature discrimination and generalization can be further improved, and these features can effectively represent people with different clothes.

## IV. EXPERIMENTS AND DISCUSSION

To evaluate the performance of our proposed SAVS framework, we perform experiments on four public cloth-changing person ReID datasets: LTCC [30], PRCC [29], Celeb-reID [27], and NKUP [42]. Since the cloth-changing person ReID task is a new and challenging research topic, to the best of our knowledge, at present, there are no comprehensive experiments with any cloth-changing ReID algorithms on all four cloth-changing person ReID datasets, and this is the first work that systematically and comprehensively assesses algorithm performance in the context of these four cloth-changing person ReID datasets. The remainder of this section is organized as follows: 1) the competitors in our experiments are listed; 2) the implementation details are described; and 3) the performance evaluations and comparisons based on these four public datasets are described.

### A. Competitors

Since the cloth-changing person ReID task is a new and challenging research topic, only a few works have been published, including SPT + ASE (TPAMI2021) [29], SE + CESD (ACCV 2020) [30], ReIDCaps (TCSVT 2020) [27], Pixel Sampling (ISPL 2021) [46], fine-grained shape-appearance mutual learning framework (FSAM) (CVPR 2021) [44], CAL [47] (CVPR 2022), AIM [36] (CVPR 2023), ACID [43] (TIP 2023), and MBUNet [35] (TIP 2023). Additionally, in the cloth-changing person ReID task, traditional person ReID algorithms are often employed, such as PCB (ECCV 2018) [9], high-order information person ReID (HoReID) (CVPR 2020) [55], MGN (ACM MM 2018) [14], ResNet50 (CVPR 2016) [53], DenseNet121 (CVPR 2017) [54], and Swin-T (ICCV 2021) [51]. Detailed information can be found in the related work.

### B. Implementation Details

Since the backbone of the SAVS approach is the Swin-T, it is also used as the baseline in our experiments. Note that the VCS module is only used in the training stage to jointly optimize the network parameters of the backbone, and in the test stage, only the backbone and the HSA module are used to extract the feature representation where the backbone only focuses on the clothing-irrelevant regions of the human. Finally, only the enhanced feature is used to describe each person in the query, where the original feature is used to reweight the enhanced feature. Specifically, an RGB image is first fed into the SCHP [50] module to obtain the human semantic information, the foreground image, and the shielding image, respectively, and then the original image, the foreground image, and the shielding image are further fed into the modules of the visual semantic decoder. Note that the enhanced feature, the original feature, and the shielding feature can be obtained in the SAVS, but only the enhanced feature is used in the query. Moreover, as the module focuses on the human semantic information and visual clothing shielding information, thus, the enhanced feature is more discriminative and robust. In addition, the default settings and divisions of these datasets [27], [29], [30], [42] are used. In our experiments, the Swin-T is first pretrained on the ImageNet dataset, and then the training samples of the LTCC, PRCC, Celeb-reID, and NKUP datasets are separately used to fine-tune the modules, including Swin-T and SAVS. In the training procedure, the minibatch size is set to 32, where each ID has four images and the input images are resized to 224 × 224. In the optimization process, stochastic gradient descent (SGD) is applied with momentum 0.9, and 60 epochs are required. Moreover, the initial learning rate is set as $3.5 \times 10^{-3}$. In addition, the learning rate is decreased by a factor of 0.1 after 40 epochs. Finally, the cumulative matching characteristics (CMCs), rank-1, and mAP are often utilized as the evaluation metrics in person ReID tasks [16], [27], [29], [46]; thus, we also strictly follow these metrics in our experiments.

### C. Performance Evaluations and Comparisons

We first assess the performances of the SAVS method when applied to four public cloth-changing person ReID datasets, and then we compare it with the abovementioned competitors. Among these approaches, if their codes could be obtained, ImageNet was first used to pretrain their backbones, and then the training samples of the LTCC, PRCC, Celeb-reID, and NKUP datasets were separately used to fine-tune the modules. Finally, the test samples of the four cloth-changing person ReID datasets were separately employed to assess their performances. If the codes were not available, the results reported by the corresponding references were used. Moreover, for a fair comparison, if the performance of a model trained by the training samples is lower than that in the corresponding reference, the results reported by the corresponding reference are also used. The results are shown in Table I. From Table I, we can obtain the following observations.

1) The SAVS consistently demonstrates exceptional performance across various datasets and approaches, yielding significant enhancements in both mAP and rank-1 values compared to state-of-the-art algorithms (with the exception of mAP on the PRCC dataset). For example, when the PRCC dataset is used, the mAP and rank-1 of the SAVS approach are 57.6% and 69.4%, respectively,

TABLE I

PERFORMANCE EVALUATION AND COMPARISON BASED ON FOUR PUBLIC CLOTH-CHANGING PERSON ReID DATASETS, WHERE THE BOLD VALUES INDICATE THE BEST PERFORMANCE IN EACH COLUMN

| Methods | Datasets | | | | | | | |
|---|---|---|---|---|---|---|---|---|
| | LTCC | | PRCC | | Celeb-reID | | NKUP | |
| | mAP | rank-1 | mAP | rank-1 | mAP | rank-1 | mAP | rank-1 |
| ResNet-50 [53] | 8.4 | 20.7 | - | 19.4 | 5.8 | 43.3 | 4.9 | 9.6 |
| DenseNet-121 [54] | 10.7 | 27.2 | 23.1 | 18.5 | 2.9 | 29.4 | 3.5 | 6.6 |
| Swin Transformer [51] | 15.2 | 42.6 | 47.6 | 46.9 | 11.1 | 46.4 | 8.5 | 16.7 |
| PCB [9] | 8.8 | 21.9 | - | 22.9 | 8.2 | 37.1 | 14.1 | 18.7 |
| MGN [14] | 10.1 | 24.2 | - | 25.9 | 10.8 | 49.0 | 16.1 | 20.6 |
| HOReID [55] | 15.1 | 46.3 | 49.2 | 37.8 | 10.2 | 48.6 | 10.0 | 16.0 |
| SE+CESD [30] | 11.7 | 25.2 | - | - | - | - | - | - |
| SPT+ASE [29] | - | - | - | 34.4 | - | - | - | - |
| ReIDCaps [27] | - | - | - | - | 15.8 | 63.0 | - | - |
| FSAM [44] | 16.2 | 38.5 | - | 54.5 | - | - | - | - |
| CAL [47] | 18.0 | 40.1 | 55.8 | 55.2 | - | - | - | - |
| Pixel Sampling [46] | 16.1 | 42.3 | 57.0 | 63.3 | 10.2 | 50.1 | 8.5 | 13.9 |
| AIM [36] | 19.1 | 40.6 | 58.3 | 57.9 | - | - | - | - |
| ACID [43] | 14.5 | 29.1 | **66.1** | 55.4 | 11.4 | 52.5 | - | - |
| MBUNet [35] | 15.0 | 40.3 | 65.2 | 68.7 | 12.8 | 55.5 | - | - |
| **SAVS** (ours) | **32.5** | **71.2** | 57.6 | **69.4** | **21.3** | **65.9** | **18.6** | **25.3** |

but the corresponding performances of the baseline are 47.6% and 46.9%, and their corresponding improvements can reach 10.0% (mAP) and 22.5% (rank-1). Similarly, the mAP and rank-1 accuracies of the SAVS approach on the LTCC dataset are 32.5% and 71.2%, respectively, but the corresponding mAP and rank-1 of the baseline can reach 15.2% and 42.6%, where the maximum improvements are 17.3% (mAP) and 28.6% (rank-1). Thus, the SAVS can significantly outperform the baseline. The reason for this is that the HSA and VCS modules are embedded into the Swin-T to extract the discriminative and robust features, and then the backbone, HSA, and VCS modules are jointly optimized. In this way, the human semantic information is fully used, and the negative effect of clothing changes is reduced as much as possible. In addition, we also observe that among these approaches, SE + CESD, SPT + ASE, ReIDCaps, FSAM, Pixel Sampling, AIM, ACID, and MBUNet are specially designed for the cloth-changing person ReID task, but their performance is still worse than that of the SAVS (with the exception of mAP on the PRCC dataset). For example, ReIDCaps has obtained good performance (second place) on the Celeb-reID dataset, where the mAP and rank-1 can reach 15.8% and 63%, respectively, but when comparing it with the SAVS, the improvements of the SAVS can achieve 5.5% (mAP) and 2.9% (rank-1), respectively. Similarly, when the Pixel Sampling method and the LTCC dataset are used, the mAP and rank-1 on the LTCC dataset are 16.1% and 42.3%, and the improvements of the SAVS can attain 16.4% and 28.9%, respectively. We also find that when the PRCC dataset is used, the mAP of the SAVS cannot always obtain the best performance, but its rank-1 can obtain the first place. The reason for this is that these specially designed methods mainly focus on body shape or contour sketch, and the complex background and human semantic information are not fully explored or are ignored, but in the SAVS, the HSA and VCS modules are designed to extract clothing-irrelevant features: more attention is given to the human semantic information, and the negative influences of the background and the cloth-changing are reduced as much as possible. Thus, the SAVS experimentally exhibits very good generalization ability, and these experimental results prove the effectiveness and robustness of the SAVS approach.

2) When comparing clothing-consistent person ReID methods with cloth-changing person ReID methods, the latter often achieve significantly better performance. For instance, the HOReID method outperforms the PCB, MGN methods on the LTCC and PRCC datasets, while the MGN method performs best on the Celeb-reID and NKUP datasets. Among the clothing-changing person ReID methods, the SAVS method stands out as the most effective. Therefore, in the following analysis, we will compare SAVS with the MGN and HOReID methods using different datasets. When evaluating the LTCC dataset, SAVS demonstrates remarkable improvements of 17.4% (mAP) and 24.9% (rank-1) compared to HOReID. Similarly, on the Celeb-reID dataset, SAVS exhibits a 10.5% improvement in mAP and a 16.9% improvement in rank-1 compared to MGN. Similar conclusions can be drawn from the remaining two datasets. The reason behind these findings is that the HOReID and MGN methods are specifically designed for clothing-consistent person ReID, assuming that a person wears the same clothes within a short time interval, resulting in similar visual appearances. However, when clothing changes occur, the person's appearance varies significantly, leading to a dramatic decline in performance. In contrast, cloth-changing person ReID methods aim to disregard clothing-related clues and focus solely on visual semantic information that remains unaffected by changes in view or posture. As a result, these methods extract clothing-irrelevant features, which greatly enhance their performance compared to the appearance-relevant features extracted by clothing-consistent person ReID methods.

3) ResNet50, DenseNet121, and Swin-T modules are widely used in many machine learning tasks, but they are also often assessed on the cloth-changing person ReID task. Although these modules can achieve good performances in many related tasks, when they are directly applied to the cloth-changing person ReID task, their performances are unsatisfactory and much worse than that of the SAVS. For example, when the PRCC dataset is used, the rank-1 accuracies of ResNet50, DenseNet121, Swin-T, and SAVS are 19.4%, 18.5%, 46.9%, and 69.4%, respectively, and the corresponding improvements achieved by the SAVS method are 50.0%, 50.9%, and 22.5%, respectively. Similarly, the mAP accuracies of ResNet50, DenseNet121, Swin-T, and SAVS on the Celeb-reID dataset are 5.8%, 2.9%, 11.1%, and 21.3%, respectively, and the corresponding improvements achieved by the SAVS method are 15.5%, 18.4%, and 10.2%, respectively. The reason for this is that although these modules are widely used in different tasks, no cloth-changing characteristics are employed in these methods, but in the SAVS, the clothing-irrelevant features are extracted, and the negative influence of the cloth-changing is reduced as much as possible. In addition, from these experiments, we can also observe that the Swin-T can achieve the best performance among the ResNet50, DenseNet121, and Swin-T modules no matter which dataset is used; thus, in our experiments, the Swin-T is also used as the backbone of the SAVS.

## V. ABLATION STUDY

An ablation study is performed using the SAVS model to analyze the contribution of each component. In this investigation, four aspects are considered: 1) the effectiveness of the HSA module; 2) the advantages of the VCS module; 3) a convergence analysis; and 4) a qualitative visualization. In the following, we discuss these four aspects separately.

### A. Effectiveness of the HSA Module

In the few existing cloth-changing person ReID methods, the global or local features are often extracted to represent the cloth-changing person, but in this section, we assess the effectiveness of the HSA module on four public cloth-changing person ReID datasets where the importance of human semantic information is discussed. Since the Swin-T is the backbone of the SAVS, in our experiments, it is also used as the baseline, where the Swin-T is used to extract features from the original image, and then the Softmax is used as the classification function. To assess the effectiveness of the original image, the foreground image, and the background image, this basic network is also used to extract the feature representations for them, and then these features are employed to find the persons from the gallery dataset. Finally, we name them "O," "F," and "B," respectively. To further evaluate the importance of human semantic information, the human semantic information (the foreground image) is fed into the HSA module, and then its outputs are used to reweight the original feature extracted by the Swin-T and the original image (we name it "O + F"). Similarly, the background image is fed into the HSA module in place of the foreground image, and the reweight operation is also used for the original feature (we call this method "O + B"). Their results are given in Table II and Fig. 4. From them, we can make the following observations.

TABLE II
EFFECTIVENESS OF THE HSA MODULE WHERE FOUR PUBLIC CLOTH-CHANGING PERSON REID DATASETS ARE EMPLOYED, AND THE BOLD VALUES INDICATE THE BEST PERFORMANCE IN EACH COLUMN. NOTE THAT $O$, $F$, AND $B$ INDICATE THE ORIGINAL IMAGE, THE FOREGROUND IMAGE, AND THE BACKGROUND IMAGE, RESPECTIVELY

| Components | | | Datasets | | | | | | | |
|---|---|---|---|---|---|---|---|---|---|---|
| | | | LTCC | | PRCC | | Celeb-reID | | NKUP | |
| $O$ | $F$ | $B$ | mAP | rank-1 | mAP | rank-1 | mAP | rank-1 | mAP | rank-1 |
| ✓ | | | 15.2 | 42.6 | 47.6 | 46.9 | 11.1 | 46.4 | 8.5 | 16.7 |
| | ✓ | | 11.7 | 34.3 | 40.0 | 45.2 | 9.8 | 40.0 | 9.0 | 14.2 |
| | | ✓ | 10.2 | 14.7 | 20.8 | 16.9 | 1.2 | 7.4 | 4.3 | 7.0 |
| ✓ | ✓ | | 17.9 | 44.3 | 50.7 | 45.2 | 9.8 | 49.7 | 10.1 | 16.1 |
| ✓ | | ✓ | **24.4** | **56.0** | **54.0** | **60.3** | **15.9** | **59.1** | **13.8** | **21.4** |

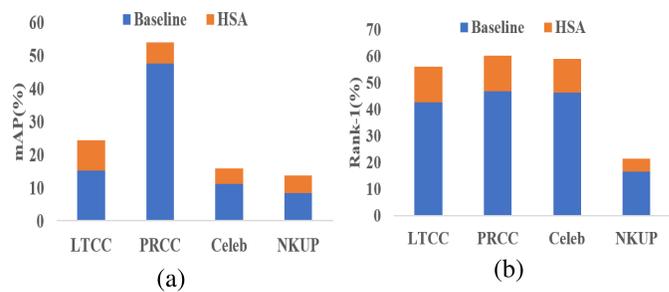

Fig. 4. Advantages of the HSA module, where four public cloth-changing datasets are utilized, and the evaluation metrics of (a) mAP and (b) rank-1 are used. Note that the blue bar indicates the results of the baseline, and the yellow bar denotes the improvements over the baseline when the HSA module is further used.

1) The original image can achieve the best performance when only the baseline is used, and the foreground image and the background image can obtain second and third places, respectively. For example, the mAP and rank-1 accuracies of the original image are 15.2% and 42.6%, respectively, on the LTCC dataset, and the corresponding accuracies of the foreground image are 11.7% and 34.3%, but the corresponding accuracies of the background image are 10.2% and 14.7%, respectively; thus, the performance of the original image can outperform the foreground and background images. Similarly, when the Celeb-reID dataset is used, the rank-1 accuracies of the original, foreground, and background images are 46.4%, 40.0%, and 7.4% and the improvement of "O" can reach 6.4% ("F") and 39% ("B"), respectively. Although person ReID is a cross-scene retrieval task and we usually assume that the background information is useless, the experimental results show that the performance of the foreground image is worse than that of the original image no matter which dataset is used. For example, when the PRCC dataset is used, the mAP accuracies of the former and latter are 47.6% and 40.0%, respectively, whose improvement can reach 7.6%. Thus, these experimental results prove that the foreground image is very useful, but the background information cannot be eliminated in its entirety.

2) The performance of "O + F" can obtain a large improvement when compared with "O" or "F" regardless of the dataset used. For example, on the LTCC dataset, the mAP accuracies of "O + F," "O," and "F" are 24.4%, 15.2%, and 11.7%, respectively, whose improvements can be 9.2% and 12.7%. When the PRCC dataset is used, the rank-1 accuracies of "O + F," "O," and "F" can reach 60.3%, 46.9%, and 45.2%, respectively, and the improvement of "O + F" can achieve 13.4% and 15.1%. The reason for this is that in "O + F," the HSA module is used to emphasize the importance of the human semantic information and reweight the visual feature map extracted by the original image. In this way, the negative effect of the background information can be reduced as much as possible, and more discriminative features can be obtained. In addition, we also observe that when the background information is fed into the HSA module, the performance of "O + B" is not stable on different datasets; for example, when the PRCC and NKUP datasets are used, the rank-1 accuracies of "O + B" are worse than those of "O," but on other datasets, the rank-1 accuracies of "O + B" can obtain little improvement. From Fig. 4, we can also obtain the same conclusions. Thus, these experiments demonstrate that the HSA module is very effective for enriching the feature representation, and the human semantic information is very helpful for feature extraction. In addition, the background information is somewhat useful for the feature representation, but emphasizing background information cannot always obtain a performance increase.

### B. Advantages of the VCS Module

To validate the advantages of the VCS module, we perform experiments on the four public cloth-changing person ReID datasets, and their results are given in Table III and Fig. 5. Note that in Table III, when the original image is fed into the Swin-T, the module is considered the baseline. Moreover, when the foreground image (human semantic) is further fed into the HSA module and its results are combined with the backbone, it is called "+HSA." Finally, when the VCS module is further used, the visual clothing shielding image is further embedded into the "+HSA" module; thus, we name it "+HSA + VCS." From them, we can see the following observations.

1) When the VCS module is used, the performance of "+HSA + VCS" can be greatly improved over that of the "+HSA" module. For example, when the Celeb-reID dataset is used, the mAP and rank-1 accuracies of "+HSA + VCS" are 21.3% and 65.9%, and the mAP and rank-1 accuracies of "+HSA" are 15.9% and 59.1%, respectively, whose improvements can reach

TABLE III
BENEFITS OF THE VCS, WHERE FOUR PUBLIC CLOTH-CHANGING DATASETS ARE UTILIZED. NOTICE THAT THE SWIN-T IS USED AS THE BASELINE, AND OUR PROPOSED HSA AND VCS MODULES ARE EMBEDDED INTO THE BASELINE STEP BY STEP

| Methods | Datasets | | | | | | | |
|---|---|---|---|---|---|---|---|---|
| | LTCC | | PRCC | | Celeb-reID | | NKUP | |
| | mAP | rank-1 | mAP | rank-1 | mAP | rank-1 | mAP | rank-1 |
| Baseline | 15.2 | 42.6 | 47.6 | 46.9 | 11.1 | 46.4 | 8.5 | 16.7 |
| +HSA | 24.4 | 56.0 | 54.0 | 60.3 | 15.9 | 59.1 | 13.8 | 21.4 |
| +HSA+VCS | 32.5 | 71.2 | 57.6 | 69.4 | 21.3 | 65.9 | 18.6 | 25.3 |

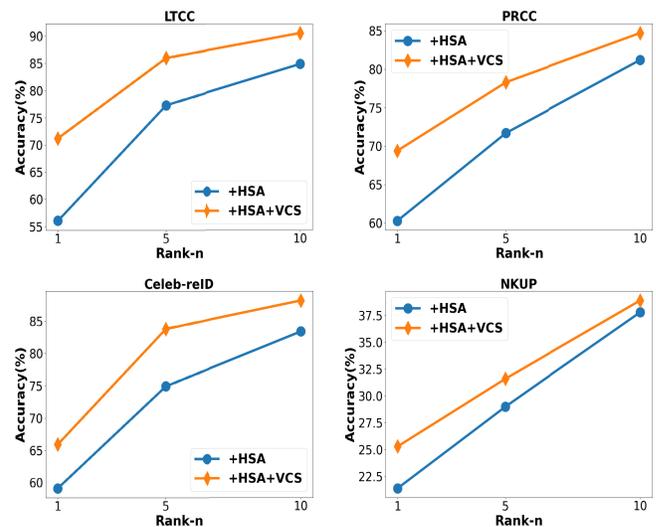

Fig. 5. Advantage analysis of the VCS module by using CMC curves on the LTCC, PRCC, Celeb-reID and NKUP datasets.

5.4% (mAP) and 6.8% (rank-1). On the PRCC dataset, the rank-1 accuracies of "+HSA + VCS" and "+HSA" are 60.3% and 69.4%, and the improvement can reach 9.1%. We can draw similar conclusions from the other datasets. In addition, when the CMC curves are used as the metric, we can also observe the same results in Fig. 5. The reason why the VCS module can be successful is that visual clothing shielding makes it difficult for appearance features to be learned, and thus, the model loses its reliance on clothing appearance when extracting features. Through the contrastive learning of the original and visual shielding features, the potential consistency can be explored to effectively solve the cloth-changing problem.

2) In the SAVS module, the original image information, visual shielding information, and human semantic information are jointly learned in a unified framework. These modules are complementary, and they can promote each other. When the HSA module and the VCS module are embedded into the baseline step by step, their combined performance can yield a stable improvement. For example, when the LTCC dataset is used, the mAP (rank-1) accuracies of the baseline, "+HSA," and "+HSA + VCS" are 15.2% (42.6%), 24.4% (56%), and 32.5% (71.2%), respectively, whose performance can be improved step by step, and the improvement of the "+HSA + VCS" can achieve 17.3% (13.4%, baseline) and 9.2% (28.6%, "+HSA"). Similarly, on the PRCC dataset, the corresponding improvements of the "+HSA + VCS" module can reach 6.4% (13.4%, baseline) and 10% (23.5%, "+HSA"), respectively.

3) In the SAVS and Pixel Sampling, both focus on the surface features of clothes; thus, we also compare their performance on two widely used public datasets, including LTCC and Celeb-reID. Since ResNet-50 is the backbone of Pixel Sampling, to obtain a fair comparison, ResNet-50 is also used as the backbone in the SAVS, but the human semantic information is ignored, and we call this approach "ResNet50 + VCS." Along another line, to assess the advantages of the backbone, the backbone of the SAVS is replaced with the Swin-T based on the "ResNet50 + VCS," and we name this approach "Swin-T + VCS." Finally, the HSA module is further assessed, and the human semantic information is embedded into "Swin-T + VCS," whose name is "Swin-T + VCS + HSA." Their results are shown in Table IV. From this, we can observe that when the same backbone and training strategy are used in the Pixel Sampling and the "ResNet50 + VCS," the performance of the latter is much better than that of the former. For example, the mAP (rank-1) accuracies of the "ResNet50 + VCS" and Pixel Sampling on the LTCC dataset are 18.7% (47.7%) and 16.1% (42.3%), respectively, whose improvement can reach 2.6% (5.4%). We can obtain the same results from the Celeb-reID dataset. The reason why the "ResNet50 + VCS" is much better than the Pixel Sampling is that the latter only focuses on a single item of clothing, but all clothing regions of the person are fully considered in the former. In this way, the negative effect of cloth-changing can be reduced as much as possible, and a more robust feature representation can be extracted by covering the clothing regions and focusing the model on the visual semantic information unrelated to the clothes. Thus, these experiments can further prove that the VCS module is very effective for solving the cloth-changing issue. In addition, when the Swin-T is further used, we can see that the performance can be further improved. For example, on the LTCC dataset, the mAP (rank-1) accuracies of "Swin-T + VCS" and Pixel Sampling are 22.8% (54.9%) and 16.1% (42.3%), respectively, whose improvement can reach 6.7% (12.6%). Thus, it can be proven that the Swin-T is also very efficient, and in our following experiments, it is utilized as the backbone of the SAVS. Finally, when the HSA module is further employed, its performance can significantly outperform the baseline, the Pixel Sampling, "ResNet50 + VCS," and "Swin-T + VCS" regardless of datasets.

### C. Convergence Analysis

In this section, we evaluate the convergence of the proposed SAVS method on four public cloth-changing person ReID datasets, including LTCC, PRCC, Celeb-reID, and NKUP, and their convergence curves are shown in Fig. 6. From Fig. 6, we can observe that the convergence speeds of the SAVS method are very fast no matter which dataset is utilized. Moreover, in the optimization process, only 30–40 epochs are required for all datasets, and the convergence curves can be stable regardless of the dataset utilized. Thus, this can further prove the effectiveness of the SAVS method.

TABLE IV
ADVANTAGE OF THE VCS AND BACKBONE ON THE LTCC AND CELEB-REID DATASETS WHEN COMPARED WITH THE PIXEL SAMPLING METHOD

| Methods | LTCC | | Celeb-reID | |
|---|---|---|---|---|
| | mAP | rank-1 | mAP | rank-1 |
| Pixel Sampling [46] | 16.1 | 42.3 | 10.2 | 50.1 |
| ResNet50+VCS | 18.7 | 47.7 | 11.6 | 52.8 |
| Swin-T+VCS | 22.8 | 54.9 | 15.2 | 57.9 |
| Swin-T+VCS+HSA | 32.5 | 71.2 | 21.3 | 65.9 |

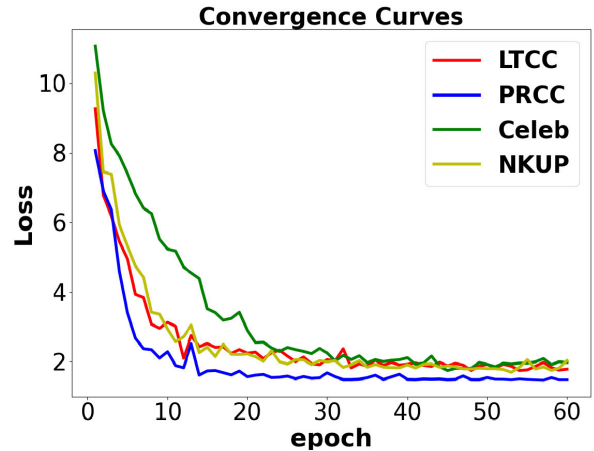

Fig. 6. Convergence curves of the SAVS method for the LTCC, PRCC, Celeb-reID, and NKUP datasets.

### D. Qualitative Visualization

To further prove the effectiveness and robustness of the SAVS, in this section, we visualize some results of the proposed SAVS on the different datasets. In this investigation, three aspects are considered: 1) visualization of the attention maps; 2) visualization of the similarity map; and 3) qualitative visualization of the retrieval results. In the following, we discuss these three aspects separately, and their results are given in Figs. 7–10. From them, we can make the following observations.

1) To further illustrate which part of the focused features are learned with the help of different modules in the SAVS, the attention maps are visualized and displayed in Fig. 7. In Fig. 7, the first row indicates the original images from different datasets. The second row ("+HSA") represents which channels on the feature map are activated by the reweighting operation where the Swin-T and the HSA module are used. The third row ("+HSA + VCS") denotes which channels between the original features and visual shielding features are more consistently constrained by the semantic alignment loss and the VCS module, where the Swin-T, the HSA, and VCS modules are utilized. From them, we can see that in the second row, the HSA module is utilized; thus, only the human semantic channels are activated where the activated regions focus on the human body, but the background information is largely ignored. Thus, these channels (corresponding to features that play a major role in later classification) are enhanced, and more discriminative and robust features can be extracted. From the third row, we can observe that the VCS module focuses on human body nonclothing regions, e.g., head, face, legs, arms, shoes, and belongings, and the human clothing regions are ignored; thus, the extracted features are clothing-irrelevant, which can effectively solve the issue of human cloth-changing, and a more discriminative and robust feature representation can be obtained. Thus, these experiments can further prove the effectiveness and advantages of the HSA and VCS modules.

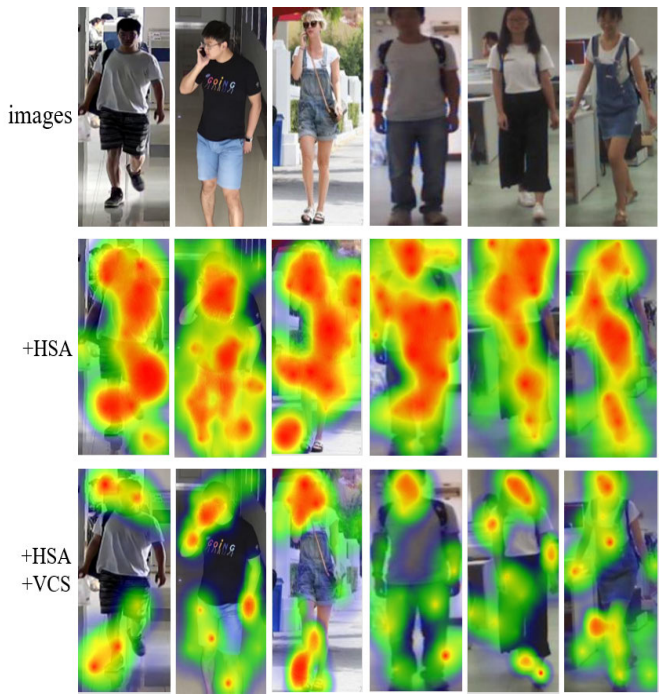

Fig. 7. Visualization of attention maps. The first row, second row, and third row indicate the original images, the results of "HSA," and the results of "+HSA + VCS," respectively. The "+HSA" focuses on the human body region, and the "+HSA + VCS" focuses on clothing-irrelevant cues, e.g., head, legs, and shoes.

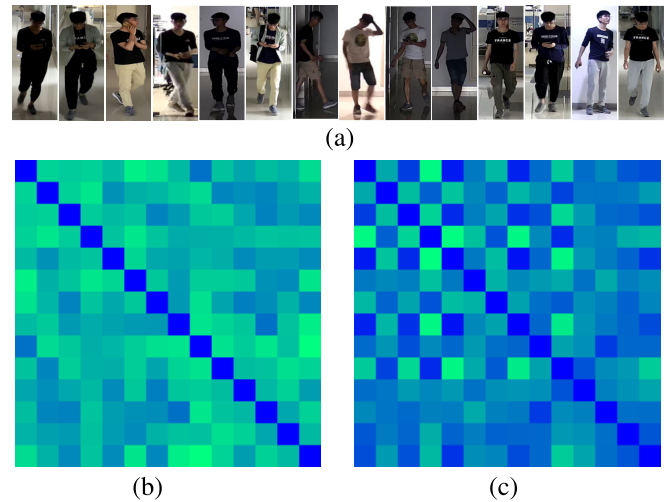

Fig. 8. Similarity maps with or without "+HSA + VCS" when the same person wears different clothes. The cosine similarity is used to calculate the distance between two images. The color of each square represents the degree of similarity between the two images indicated by the horizontal and vertical coordinates. The blue and green colors represent the most and the least similar pairs, respectively. (a) 14 images of the same person wearing different clothes. (b) Similarity map without the HSA and VCS modules. (c) Similarity map with the HSA and VCS modules.

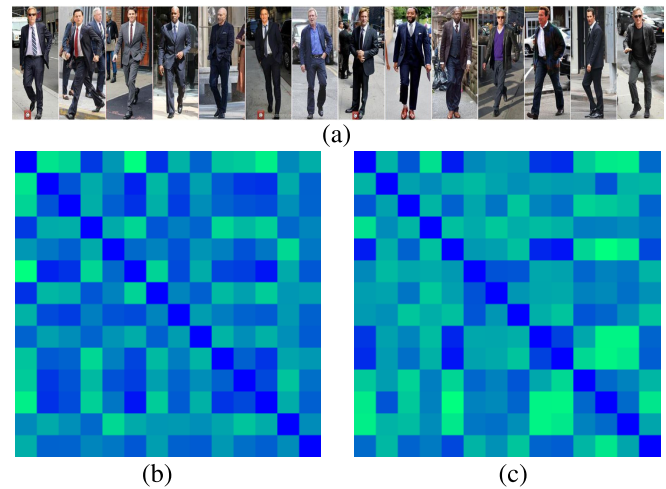

Fig. 9. Similarity maps with or without "+HSA + VCS" when different persons wear similar clothes. The cosine similarity is used to calculate the distance between two images. The color of each square represents the degree of similarity between the two images indicated by the horizontal and vertical coordinates. The blue and green colors represent the most and the least similar pairs, respectively. (a) 14 images of the different person wearing similar clothes. (b) Similarity map without the HSA and VCS modules. (c) Similarity map with the HSA and VCS modules.

2) To intuitively illustrate the effectiveness of the HSA and VCS modules from another view, the feature similarities between different images are calculated. Specifically, 14 images of the same person wearing different clothes are first selected, and then the features are extracted for each image by the Swin-T (our baseline). Moreover, the cosine similarities between any two images are calculated by the corresponding features. Finally, we repeat the above operation in pairs for all 14 images and visualize their similarities to obtain a 14 × 14 similarity map. Similarly, we can also extract these image features by Swin-T + HSA + VCS (our proposed SAVS module), and then the cosine similarities between them and the similarity map are calculated. The results are given in Fig. 8, where the blue and green colors represent the most and the least similar pairs, respectively. From this, we can determine that when only the baseline is utilized, the extracted features are very relative to the clothes, and the features are not discriminative and robust; thus, the similarities between the same person wearing different clothes are very small (the color of most of the squares is green). However, when the HSA and VCS modules are embedded into the baseline, the human semantic information and the clothing-irrelevant clues are fully explored; thus, the extracted features are more discriminative and robust, where similarities between the same person wearing different clothes are added, and the color of most of the squares is blue. We also can observe the similar conclusions from Fig. 9 where different persons wear similar clothes. Thus, these experiments can further prove that the SAVS is very effective, efficient, and robust.

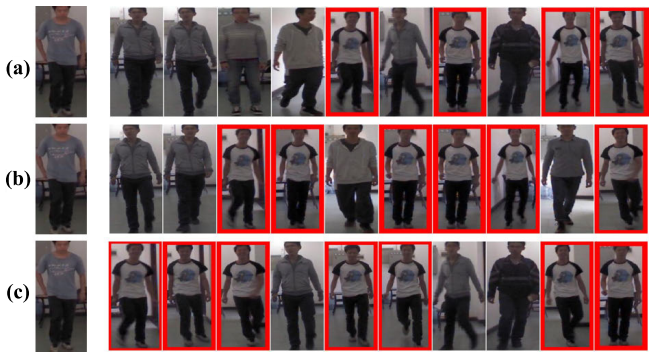

Fig. 10. Qualitative visualization of (a) baseline, (b) baseline + HSA, and (c) baseline + HSA + VCS (SAVS) on the PRCC dataset. The top-left column is a single query image, and the other columns represent the top ten retrieval results. Note that the red boxes indicate the correct results.

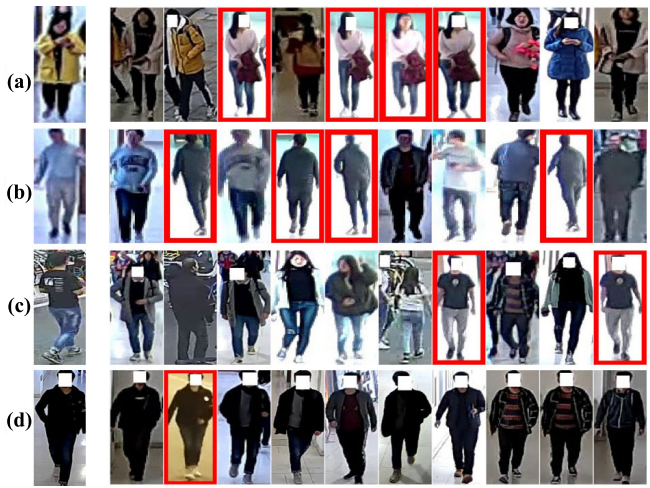

Fig. 11. Qualitative visualization of the SAVS on the NKUP dataset. The top-left column is a single query image, and the other columns represent the top ten retrieval results. Note that the red boxes indicate the correct results. (a) and (b) Full-frontal images are given. (c) and (d) Provided image is a back image or the face is obscured.

3) To further prove the effectiveness and robustness of the SAVS method, in this section, the retrieval results of the SAVS on the PRCC and NKUP datasets are given in Figs. 10 and 11, respectively, where each row is a retrieval example including one query image and the top ten most similar images. We notice that in Fig. 10(a)–(c), the effectiveness of different parts of the SAVS, including the baseline, the baseline + HSA, and the baseline + HSA + VCS, is visualized with respect to the PRCC dataset. Fig. 10 demonstrates that since the visual appearance of the cloth-changing person drastically changes, the baseline still has difficulty obtaining the correct retrieval results when only the original feature is extracted. When the human semantic information is further combined, the feature discrimination is improved. Although the number of correct retrieval results is increased significantly, the first correct retrieval result lies in the third position, and the average performance is not unsatisfactory. Finally, when visual shielding information is employed, more correct retrieval results can be returned; moreover, their locations are very close to the front of the images. Thus, these results can further prove that the HSA and VCS modules are very useful for describing cloth-changing persons, and the extracted feature is effective and robust. To further demonstrate the effectiveness of the SAVS modules on difficult samples, we visualize four retrieval results of the SAVS module on the NKUP dataset, where the face information is often covered, and the results are presented in Fig. 11. In the two retrieval cases [see Fig. 11(a) and (b)], the full-frontal image is given, but in the gallery, the facial information of one person is often obscured in Fig. 11(a), and the person is captured with mainly back profile information in Fig. 11(b). Luckily, in this case, the SAVS can still retrieve 4–5 correct results effectively, and these results are also at the top of the returned results. In both retrieval cases [see Fig. 11(c) and (d)], where the provided image is a back image or the face is obscured, the SAVS can still correctly identify persons, but the average result declines dramatically, where only one or two correct results are returned. These experimental results indicate that it is quite challenging to perform cloth-changing person ReID when the provided image lacks visual semantics to a large extent.

## VI. CONCLUSION

This work proposed a novel SAVS algorithm for the cloth-changing person ReID task, where the key idea is to shield clues related to the appearance of clothes and only focus on visual semantic information that is not sensitive to view/posture changes. In the SAVS, an HSA module is designed to highlight human information and reweight the visual feature map, and a VCS module is proposed to extract a more robust feature representation for cloth-changing tasks by covering clothing regions and focusing the model on visual semantic information unrelated to clothes. Most importantly, these two modules are jointly explored in an end-to-end unified framework. The results of extensive experiments conducted on four cloth-changing person ReID datasets demonstrate that the SAVS can significantly outperform state-of-the-art cloth-changing person ReID methods in terms of both mAP and rank-1 accuracy, and more discriminative and robust features can be extracted to represent the cloth-changing persons. An ablation study also proves that human semantic information and visual shielding information are very helpful for solving the cloth-changing person ReID task where clothes-independent features can be obtained. Additionally, different qualitative visualizations can further prove the effectiveness and robustness of the HSA and VCS modules.

In the future, we intend to explore how to shield the effects of clothing changes and design approaches based on novel dynamic features such as gait that can also describe static person images.

## REFERENCES

[1] H. Zhang et al., "Attribute-guided collaborative learning for partial person re-identification," *IEEE Trans. Pattern Anal. Mach. Intell.*, early access, Sep. 5, 2023, doi: 10.1109/TPAMI.2023.3312302.
[2] R. Hou, B. Ma, H. Chang, X. Gu, S. Shan, and X. Chen, "IAUnet: Global context-aware feature learning for person reidentification," *IEEE Trans. Neural Netw. Learn. Syst.*, vol. 32, no. 10, pp. 4460–4474, Oct. 2021.
[3] Z. Zheng, L. Zheng, and Y. Yang, "Pedestrian alignment network for large-scale person re-identification," *IEEE Trans. Circuits Syst. Video Technol.*, vol. 29, no. 10, pp. 3037–3045, Oct. 2019.
[4] R. Zhou, X. Chang, L. Shi, Y.-D. Shen, Y. Yang, and F. Nie, "Person reidentification via multi-feature fusion with adaptive graph learning," *IEEE Trans. Neural Netw. Learn. Syst.*, vol. 31, no. 5, pp. 1592–1601, May 2020.

[5] L. Gao, H. Zhang, Z. Gao, W. Guan, Z. Cheng, and M. Wang, "Texture semantically aligned with visibility-aware for partial person re-identification," in *Proc. 28th ACM Int. Conf. Multimedia*, Oct. 2020, pp. 3771–3779.

[6] Y. Lin et al., "Improving person re-identification by attribute and identity learning," *Pattern Recognit.*, vol. 95, pp. 151–161, Nov. 2019.

[7] J. Yang, W.-S. Zheng, Q. Yang, Y.-C. Chen, and Q. Tian, "Spatial-temporal graph convolutional network for video-based person re-identification," in *Proc. IEEE/CVF Conf. Comput. Vis. Pattern Recognit. (CVPR)*, Jun. 2020, pp. 3286–3296.

[8] L. Wei, S. Zhang, H. Yao, W. Gao, and Q. Tian, "GLAD: Global–local-alignment descriptor for scalable person re-identification," *IEEE Trans. Multimedia*, vol. 21, no. 4, pp. 986–999, Apr. 2019.

[9] Y. Sun, L. Zheng, Y. Yang, Q. Tian, and S. Wang, "Beyond part models: Person retrieval with refined part pooling (and a strong convolutional baseline)," in *Proc. Eur. Conf. Comput. Vis. (ECCV)*, 2018, pp. 480–496.

[10] B. Sun, Y. Ren, and X. Lu, "Semisupervised consistent projection metric learning for person reidentification," *IEEE Trans. Cybern.*, vol. 52, no. 2, pp. 738–747, Feb. 2022.

[11] Y. Sun, L. Zheng, Y. Li, Y. Yang, Q. Tian, and S. Wang, "Learning part-based convolutional features for person re-identification," *IEEE Trans. Pattern Anal. Mach. Intell.*, vol. 43, no. 3, pp. 902–917, Mar. 2021.

[12] A. J. Ma, J. Li, P. C. Yuen, and P. Li, "Cross-domain person reidentification using domain adaptation ranking SVMs," *IEEE Trans. Image Process.*, vol. 24, no. 5, pp. 1599–1613, May 2015.

[13] Z. Liu, H. Lu, X. Ruan, and M.-H. Yang, "Person reidentification by joint local distance metric and feature transformation," *IEEE Trans. Neural Netw. Learn. Syst.*, vol. 30, no. 10, pp. 2999–3009, Oct. 2019.

[14] G. Wang, Y. Yuan, X. Chen, J. Li, and X. Zhou, "Learning discriminative features with multiple granularities for person re-identification," in *Proc. 26th ACM Int. Conf. Multimedia*, Oct. 2018, pp. 274–282.

[15] H. Wang, L. Jiao, S. Yang, L. Li, and Z. Wang, "Simple and effective: Spatial Rescaling for person reidentification," *IEEE Trans. Neural Netw. Learn. Syst.*, vol. 33, no. 1, pp. 145–156, Jan. 2022.

[16] Z. Gao, L. Gao, H. Zhang, Z. Cheng, R. Hong, and S. Chen, "DCR: A unified framework for holistic/partial person ReID," *IEEE Trans. Multimedia*, vol. 23, pp. 3332–3345, 2021.

[17] M. Ye, J. Shen, G. Lin, T. Xiang, L. Shao, and S. C. H. Hoi, "Deep learning for person re-identification: A survey and outlook," *IEEE Trans. Pattern Anal. Mach. Intell.*, vol. 44, no. 6, pp. 2872–2893, Jun. 2022.

[18] W. Zhang, Y. Chen, W. Yang, G. Wang, J.-H. Xue, and Q. Liao, "Class-variant margin normalized softmax loss for deep face recognition," *IEEE Trans. Neural Netw. Learn. Syst.*, vol. 32, no. 10, pp. 4742–4747, Oct. 2021.

[19] D. Liu, X. Gao, C. Peng, N. Wang, and J. Li, "Heterogeneous face interpretable disentangled representation for joint face recognition and synthesis," *IEEE Trans. Neural Netw. Learn. Syst.*, vol. 33, no. 10, pp. 5611–5625, Oct. 2022.

[20] X. Luo, H. Wu, Z. Wang, J. Wang, and D. Meng, "A novel approach to large-scale dynamically weighted directed network representation," *IEEE Trans. Pattern Anal. Mach. Intell.*, vol. 44, no. 12, pp. 9756–9773, Dec. 2022.

[21] J. Wang, L. Wang, F. Nie, and X. Li, "Joint feature selection and extraction with sparse unsupervised projection," *IEEE Trans. Neural Netw. Learn. Syst.*, vol. 34, no. 6, pp. 3071–3081, Jun. 2021.

[22] X. Luo, H. Wu, and Z. Li, "NeuLFT: A novel approach to nonlinear Canonical Polyadic decomposition on high-dimensional incomplete tensors," *IEEE Trans. Knowl. Data Eng.*, vol. 35, no. 6, pp. 6148–6166, Jun. 2023.

[23] L. Zheng, L. Shen, L. Tian, S. Wang, J. Wang, and Q. Tian, "Scalable person re-identification: A benchmark," in *Proc. IEEE Int. Conf. Comput. Vis. (ICCV)*, Dec. 2015, pp. 1116–1124.

[24] Z. Zheng, L. Zheng, and Y. Yang, "Unlabeled samples generated by GAN improve the person re-identification baseline in vitro," in *Proc. IEEE Int. Conf. Comput. Vis. (ICCV)*, Oct. 2017, pp. 3774–3782.

[25] W. Li, R. Zhao, T. Xiao, and X. Wang, "DeepReID: Deep filter pairing neural network for person re-identification," in *Proc. IEEE Conf. Comput. Vis. Pattern Recognit.*, Jun. 2014, pp. 152–159.

[26] S. Li et al., "Cocas : Large-scale clothes-changing person re-identification with clothes templates," *IEEE Trans. Circuits Syst. Video Technol.*, vol. 33, no. 4, pp. 1839–1853, Apr. 2023.

[27] Y. Huang, J. Xu, Q. Wu, Y. Zhong, P. Zhang, and Z. Zhang, "Beyond scalar neuron: Adopting vector-neuron capsules for long-term person re-identification," *IEEE Trans. Circuits Syst. Video Technol.*, vol. 30, no. 10, pp. 3459–3471, Oct. 2020.

[28] Z. Cui, J. Zhou, Y. Peng, S. Zhang, and Y. Wang, "DCR-ReID: Deep component reconstruction for cloth-changing person re-identification," *IEEE Trans. Circuits Syst. Video Technol.*, vol. 33, no. 8, pp. 4415–4428, Aug. 2023.

[29] Q. Yang, A. Wu, and W.-S. Zheng, "Person re-identification by contour sketch under moderate clothing change," *IEEE Trans. Pattern Anal. Mach. Intell.*, vol. 43, no. 6, pp. 2029–2046, Jun. 2021.

[30] X. Qian et al., "Long-term cloth-changing person re-identification," in *Proc. 15th Asian Conf. Comput. Vis.*, vol. 12624, 2020, pp. 71–88.

[31] Z. Zheng, X. Yang, Z. Yu, L. Zheng, Y. Yang, and J. Kautz, "Joint discriminative and generative learning for person re-identification," in *Proc. IEEE/CVF Conf. Comput. Vis. Pattern Recognit. (CVPR)*, Jun. 2019, pp. 2133–2142.

[32] S. Yu, S. Li, D. Chen, R. Zhao, J. Yan, and Y. Qiao, "COCAS: A large-scale clothes changing person dataset for re-identification," in *Proc. IEEE/CVF Conf. Comput. Vis. Pattern Recognit. (CVPR)*, Jun. 2020, pp. 3397–3406.

[33] X. Jin et al., "Cloth-changing person re-identification from a single image with gait prediction and regularization," in *Proc. IEEE/CVF Conf. Comput. Vis. Pattern Recognit. (CVPR)*, New Orleans, LA, USA, Jun. 2022, pp. 14258–14267.

[34] S. Yang, B. Kang, and Y. Lee, "Sampling agnostic feature representation for long-term person re-identification," *IEEE Trans. Image Process.*, vol. 31, pp. 6412–6423, 2022.

[35] G. Zhang, J. Liu, Y. Chen, Y. Zheng, and H. Zhang, "Multi-biometric unified network for cloth-changing person re-identification," *IEEE Trans. Image Process.*, vol. 32, pp. 4555–4566, 2023.

[36] Z. Yang, M. Lin, X. Zhong, Y. Wu, and Z. Wang, "Good is bad: Causality inspired cloth-debiasing for cloth-changing person re-identification," in *Proc. IEEE/CVF Conf. Comput. Vis. Pattern Recognit. (CVPR)*, Vancouver, BC, Canada, Jun. 2023, pp. 1472–1481.

[37] S. Paisitkriangkrai, C. Shen, and A. van den Hengel, "Learning to rank in person re-identification with metric ensembles," in *Proc. IEEE Conf. Comput. Vis. Pattern Recognit. (CVPR)*, Jun. 2015, pp. 1846–1855.

[38] Y. Shen, T. Xiao, H. Li, S. Yi, and X. Wang, "End-to-end deep Kronecker-product matching for person re-identification," in *Proc. IEEE/CVF Conf. Comput. Vis. Pattern Recognit.*, Jun. 2018, pp. 6886–6895.

[39] C. Song, Y. Huang, W. Ouyang, and L. Wang, "Mask-guided contrastive attention model for person re-identification," in *Proc. IEEE/CVF Conf. Comput. Vis. Pattern Recognit.*, Jun. 2018, pp. 1179–1188.

[40] J. Miao, Y. Wu, P. Liu, Y. Ding, and Y. Yang, "Pose-guided feature alignment for occluded person re-identification," in *Proc. IEEE/CVF Int. Conf. Comput. Vis. (ICCV)*, Oct. 2019, pp. 542–551.

[41] Z. Zheng, X. Wang, N. Zheng, and Y. Yang, "Parameter-efficient person re-identification in the 3D space," *IEEE Trans. Neural Netw. Learn. Syst.*, early access, Oct. 31, 2022, doi: 10.1109/TNNLS.2022.3214834.

[42] K. Wang, Z. Ma, S. Chen, J. Yang, K. Zhou, and T. Li, "A benchmark for clothes variation in person re-identification," *Int. J. Intell. Syst.*, vol. 35, no. 12, pp. 1881–1898, Dec. 2020.

[43] Z. Yang, X. Zhong, Z. Zhong, H. Liu, Z. Wang, and S. Satoh, "Win-win by competition: Auxiliary-free cloth-changing person re-identification," *IEEE Trans. Image Process.*, vol. 32, pp. 2985–2999, 2023.

[44] P. Hong, T. Wu, A. Wu, X. Han, and W.-S. Zheng, "Fine-grained shape-appearance mutual learning for cloth-changing person re-identification," in *Proc. IEEE/CVF Conf. Comput. Vis. Pattern Recognit. (CVPR)*, Jun. 2021, pp. 10508–10517.

[45] Z. Gao, H. Wei, W. Guan, W. Nie, M. Liu, and M. Wang, "Multigranular visual-semantic embedding for cloth-changing person re-identification," in *Proc. 30th ACM Int. Conf. Multimedia*, Lisboa, Portugal, Oct. 2022, pp. 3703–3711.

[46] X. Shu, G. Li, X. Wang, W. Ruan, and Q. Tian, "Semantic-guided pixel sampling for cloth-changing person re-identification," *IEEE Signal Process. Lett.*, vol. 28, pp. 1365–1369, 2021.

[47] X. Gu, H. Chang, B. Ma, S. Bai, S. Shan, and X. Chen, "Clothes-changing person re-identification with RGB modality only," in *Proc. IEEE/CVF Conf. Comput. Vis. Pattern Recognit. (CVPR)*, Jun. 2022, pp. 1050–1059.

[48] Z. Liu et al., "Towards natural and accurate future motion prediction of humans and animals," in *Proc. IEEE/CVF Conf. Comput. Vis. Pattern Recognit. (CVPR)*, Jun. 2019, pp. 9996–10004.

[49] Z. Gao, L. Guo, W. Guan, A.-A. Liu, T. Ren, and S. Chen, "A pairwise attentive adversarial spatiotemporal network for cross-domain few-shot action recognition-R2," *IEEE Trans. Image Process.*, vol. 30, pp. 767–782, 2021.

[50] P. Li, Y. Xu, Y. Wei, and Y. Yang, "Self-correction for human parsing," *IEEE Trans. Pattern Anal. Mach. Intell.*, vol. 44, no. 6, pp. 3260–3271, Jun. 2022.
[51] Z. Liu et al., "Swin Transformer: Hierarchical vision transformer using shifted windows," in *Proc. IEEE/CVF Int. Conf. Comput. Vis. (ICCV)*, Oct. 2021, pp. 9992–10002.
[52] Y. Sun et al., "Circle loss: A unified perspective of pair similarity optimization," in *Proc. IEEE/CVF Conf. Comput. Vis. Pattern Recognit. (CVPR)*, Jun. 2020, pp. 6397–6406.
[53] K. He, X. Zhang, S. Ren, and J. Sun, "Deep residual learning for image recognition," in *Proc. IEEE Conf. Comput. Vis. Pattern Recognit. (CVPR)*, Jun. 2016, pp. 770–778.
[54] G. Huang, Z. Liu, L. Van Der Maaten, and K. Q. Weinberger, "Densely connected convolutional networks are similar to traditional convolutional networks, but they have more layers," in *Proc. IEEE Conf. Comput. Vis. Pattern Recognit. (CVPR)*, Jul. 2017, pp. 2261–2269.
[55] G. Wang et al., "High-order information matters: Learning relation and topology for occluded person re-identification," in *Proc. IEEE/CVF Conf. Comput. Vis. Pattern Recognit. (CVPR)*, Jun. 2020, pp. 6448–6457.

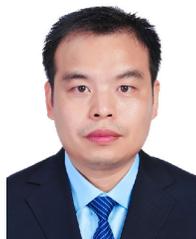
**Zan Gao** (Member, IEEE) received the Ph.D. degree from the Beijing University of Posts and Telecommunications, Beijing, China, in 2011.

From September 2009 to September 2010, he was with the School of Computer Science, Carnegie Mellon University, Pittsburgh, PA, USA. From July 2016 to January 2017, he was with the School of Computing, National University of Singapore, Singapore. He is currently a Full Professor with the Shandong Artificial Intelligence Institute, Qilu University of Technology (Shandong Academy of Sciences), Jinan, China. He has authored over 80 scientific papers in international conferences and journals, including IEEE TRANSACTIONS ON PATTERN ANALYSIS AND MACHINE INTELLIGENCE (TPAMI), IEEE TRANSACTIONS ON IMAGE PROCESSING (TIP), IEEE TRANSACTIONS ON NEURAL NETWORKS AND LEARNING SYSTEMS (TNNLS), IEEE TRANSACTIONS ON MULTIMEDIA (TMM), IEEE TRANSACTIONS ON CYBERNETICS (TCYBE), ACM Transactions on Multimedia Computing, Communications, and Applications (TOMM), IEEE Conference on Computer Vision and Pattern Recognition (CVPR), ACM International Conference on Multimedia (ACM MM), International World Wide Web Conference (WWW), International Conference on Research and Development in Information Retrieval (SIGIR), Association for the Advancement of Artificial Intelligence (AAAI), *Neural Networks*, and *Internet of Things*. His research interests include artificial intelligence, multimedia analysis and retrieval, and machine learning.

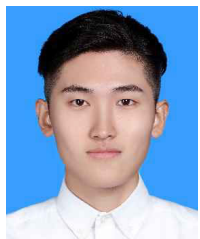
**Hongwei Wei** is currently pursuing the master's degree with the Shandong Artificial Intelligence Institute, Qilu University of Technology (Shandong Academy of Sciences), Jinan, China.

His research interests include artificial intelligence, multimedia analysis and retrieval, computer vision, and machine learning

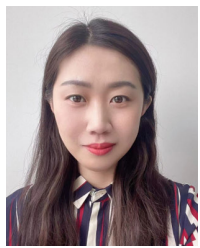
**Weili Guan** (Member, IEEE) received the bachelor's degree from Huaqiao University, Quanzhou, China, in 2009, and the Graduate Diploma and master's degrees from the National University of Singapore, Singapore, in 2011 and 2014, respectively. She is currently pursuing the Ph.D. degree with the Faculty of Information Technology, Monash University, Clayton, VIC, Australia.

She joined Hewlett Packard Enterprise, Singapore, as a Software Engineer, and worked there for around five years. Her research interests include multimedia computing and information retrieval.

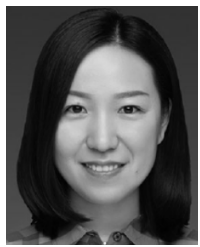
**Jie Nie** (Member, IEEE) received the Ph.D. degree in computer science from the Ocean University of China, Qingdao, China, in 2011.

From September 2009 to September 2010, she was a Visiting Scholar with the School of Electrical Engineering, University of Pittsburgh, Pittsburgh, PA, USA. She held a post-doctoral position with Tsinghua University, Beijing, China, from 2015 to 2017. She is currently with the Ocean University of China. Her current research interests include social media and multimedia content analysis.

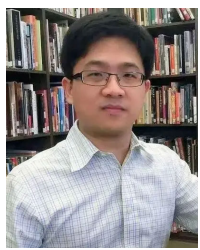
**Meng Wang** (Fellow, IEEE) received the B.E. and Ph.D. degrees from the Special Class for the Gifted Young and the Department of Electronic Engineering and Information Science, University of Science and Technology of China (USTC), Hefei, China, in 2003 and 2008, respectively.

He is currently a Professor with the Hefei University of Technology, Hefei. His current research interests include multimedia content analysis, search, mining, recommendation, and large-scale computing.

Dr. Wang received the Best Paper Awards successively from the 17th and 18th ACM International Conference on Multimedia, the Best Paper Award from the 16th International Multimedia Modeling Conference, the Best Paper Award from the 4th International Conference on Internet Multimedia Computing and Service, and the Best Demo Award from the 20th ACM International Conference on Multimedia.

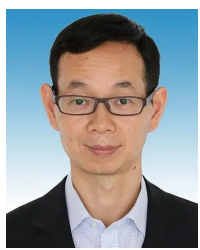
**Shengyong Chen** (Senior Member, IEEE) received the Ph.D. degree in computer vision from the City University of Hong Kong, Hong Kong, in 2003.

He was with the University of Hamburg, Hamburg, Germany, from 2006 to 2007. He is currently a Professor with the Tianjin University of Technology, Tianjin, China, and with the Zhejiang University of Technology, Hangzhou, China. He has authored over 100 scientific articles in international journals. His research interests include computer vision, robotics, and image analysis.

Dr. Chen is a fellow of the IET and a Senior Member of the CCF. He received the National Outstanding Youth Foundation Award of China in 2013. He also received the fellowship from the Alexander von Humboldt Foundation of Germany.